\documentclass[twoside,11pt]{article}
\usepackage[preprint]{jmlr2e}

\ShortHeadings{Tensor Logic}{Domingos}
\firstpageno{1}

\begin{document}

\title{Tensor Logic: The Language of AI}

\author{\name Pedro Domingos \email pedrod@cs.washington.edu \\
\addr Paul G. Allen School of Computer Science \& Engineering \\
University of Washington \\
Seattle, WA 98195-2350, USA }

\maketitle

\begin{abstract}
Progress in AI is hindered by the lack of a programming language with all the requisite features. Libraries like PyTorch and TensorFlow provide automatic differentiation and efficient GPU implementation, but are additions to Python, which was never intended for AI. Their lack of support for automated reasoning and knowledge acquisition has led to a long and costly series of hacky attempts to tack them on. On the other hand, AI languages like LISP and Prolog lack scalability and support for learning. This paper proposes tensor logic, a language that solves these problems by unifying neural and symbolic AI at a fundamental level. The sole construct in tensor logic is the tensor equation, based on the observation that logical rules and Einstein summation are essentially the same operation, and all else can be reduced to them. I show how to elegantly implement key forms of neural, symbolic and statistical AI in tensor logic, including transformers, formal reasoning, kernel machines and graphical models. Most importantly, tensor logic makes new directions possible, such as sound reasoning in embedding space. This combines the scalability and learnability of neural networks with the reliability and transparency of symbolic reasoning, and is potentially a basis for the wider adoption of AI.
\end{abstract}

\begin{keywords}
\hspace{-0.1in}
deep learning, automated reasoning, knowledge representation, logic programming, Einstein summation, embeddings, kernel machines, probabilistic graphical models
\end{keywords}

\section{Introduction}

Fields take off when they find their language. Physics took off when Newton invented calculus, and couldn't have done so without it. Maxwell's equations would be unusable without Heaviside's vector calculus notation. As mathematicians and physicists like to say, a good notation is half the battle. Much of electrical engineering would be impossible without complex numbers, and digital circuits without Boolean logic. Modern chip design is made possible by hardware description languages, databases by relational algebra, the Internet by the Internet Protocol, and the Web by HTML. More generally, computer science would not have gotten far without high-level programming languages. Qualitative fields also depend critically on their terminology. Even artists rely on the idioms and stylistic conventions of their genre for their work.

A field's language saves its practitioners time, focuses their attention, and changes how they think. It unites the field around common directions and decreases entropy. It makes key things obvious and avoids repeatedly hacking solutions from scratch.

Has AI found its language? LISP, one of the first high-level programming languages, made symbolic AI possible. In the 80s Prolog also became popular. Both, however, suffered from poor scalability and lack of support for learning, and were ultimately displaced, even within AI, by general-purpose languages like Java and C{\small ++}. Graphical models provide a {\em lingua franca} for probabilistic AI, but their applicability is limited by the cost of inference. Formalisms like Markov logic seamlessly combine symbolic and probabilistic AI, but are also hindered by the cost of inference.

Python is currently the {\em de facto} language of AI, but was never designed for it, and it shows. Libraries like PyTorch and TensorFlow provide important features like automatic differentiation and GPU implementation, but are of no help for key tasks like automated reasoning and knowledge acquisition. Neurosymbolic AI seeks to ameliorate this by combining deep learning modules with symbolic AI ones, but often winds up having the shortcomings of both. In sum, AI has clearly not found its language yet.

There are clear desiderata for such a language. Unlike Python, it should hide everything that is not AI, allowing AI programmers to focus on what matters. It should facilitate incorporating prior knowledge into AI systems and reasoning automatically over it. It should also facilitate learning automatically, and the resulting models should be transparent and reliable. It should scale effortlessly. Symbolic AI has some of these properties and deep learning has others, but neither has all. We therefore need to merge them.

Tensor logic does this by unifying their mathematical foundations. It is based on the observation that essentially all neural networks can be constructed using tensor algebra, all symbolic AI using logic programming, and the two are fundamentally equivalent, differing only in the atomic data types used.

I begin with a brief review of logic programming and tensor algebra. The core of the paper defines tensor logic and describes its inference and learning engines. I then show how to elegantly implement neural networks, symbolic AI, kernel machines and graphical models in it. I show how tensor logic enables reliable and transparent reasoning in embedding space. I propose two approaches to scaling it up. The paper concludes with a discussion of other potential uses of tensor logic, prospects for its wide adoption, and next steps toward it.

\section{Background}

\subsection{Logic Programming}

The most widely used formalism in symbolic AI is logic programming \citep{lloyd87}. The simplest logic programming language, which suffices for our purposes, is Datalog \citep{greco16}. A Datalog program is a set of {\em rules} and {\em facts}. A fact is a statement of the form $r(o_1, \ldots, o_n)$, where $r$ is a relation name and the $o$'s are object names. For example, ${\tt Parent(Bob, Charlie)}$ states that Bob is a parent of Charlie, and ${\tt Ancestor(Alice, Bob)}$ that Alice is an ancestor of Bob. A rule is a statement of the form $A_0 \leftarrow A_1, \ldots, A_m$, where the arrow means ``if'', commas denote conjunction, and each of the $A$'s has the form $r(x_1, \ldots, x_n)$, with $r$ being a relation name and the $x$'s being variables or object names. For example, the rule
\[ {\tt Ancestor(x, y)} \leftarrow {\tt Parent(x, y)} \]
says that parents are ancestors, and the rule
\[ {\tt Ancestor(x, z)} \leftarrow {\tt Ancestor(x, y)}, {\tt Parent(y, z)} \]
says that ${\tt x}$ is ${\tt z}$'s ancestor if ${\tt x}$ is ${\tt y}$'s ancestor and ${\tt y}$ is ${\tt z}$'s parent. Informally, a rule says that its left-hand side or {\em head} is true if there are known facts that make all the relations on its right-hand side or {\em body} simultaneously true. For example, the rules and facts above imply that ${\tt Ancestor(Alice, Charlie)}$ is true.

In database terminology, a Datalog rule is a series of {\em joins} followed by a {\em projection}. The (natural) join of two relations $R$ and $S$ is the set of all tuples that can be formed from tuples in $R$ and $S$ having the same values for the same arguments. When two relations have no arguments in common, their join reduces to their Cartesian product. The projection of a relation $R$ onto a subset $G$ of its arguments is the relation obtained by discarding from the tuples in $R$ all arguments not in $G$. For example, the rule
\[ {\tt Ancestor(x, z)} \leftarrow {\tt Ancestor(x, y)}, {\tt Parent(y, z)} \]
joins the relations ${\tt Ancestor(x, y)}$ and ${\tt Parent(y, z)}$ on ${\tt y}$ and projects the result onto $\{ {\tt x}, {\tt z} \}$; the tuples ${\tt Ancestor(Alice, Bob)}$ and ${\tt Parent(Bob, Charlie)}$ yield the tuple ${\tt Ancestor(Alice,}$ ${\tt Charlie)}$.

Two common inference algorithms in logic programming are forward and backward chaining. In forward chaining, the rules are repeatedly applied to the known facts to derive new facts until no further ones can be derived. The result is called the {\em deductive closure} or {\em fixpoint} of the program, and all questions of interest can be answered simply by examining it. For example, the answer to the query ${\tt Ancestor(Alice, x)}$ (``Who is Alice an ancestor of?'') given the rules and facts above is $\{ {\tt Bob}, {\tt Charlie} \}$.

Backward chaining attempts to answer a question by finding facts that match it or rules that have it as their head and facts that match the body, and so on recursively. For example, the query ${\tt Ancestor(Alice, Charlie)}$ does not match any facts, but it matches the rule
\[ {\tt Ancestor(x, z)} \leftarrow {\tt Ancestor(x, y)}, {\tt Parent(y, z)} \]
and this rule's body matches the facts ${\tt Ancestor(Alice, Bob)}$ and ${\tt Parent(Bob, Charlie)}$, and therefore the answer is ${\tt True}$.

Forward and backward chaining in Datalog are {\em sound} inference procedures, meaning that the answers they give are guaranteed to follow logically from the rules and facts in the program. Logic programs have both {\em declarative} and {\em procedural} semantics, meaning a rule can be interpreted both as a statement about the world and as a procedure for computing its head with the given arguments by calling the procedures in the body and combining the results.

The field of inductive logic programming (ILP) is concerned with learning logic programs from data \citep{lavrac94}. For example, an ILP system might induce the rules above from a small database of parent and ancestor relations. Once induced, these rules can answer questions about ancestry chains of any length and involving anyone. Some ILP systems can also do {\em predicate invention}, i.e., discover relations that do not appear explicitly in the data, akin to hidden variables in neural networks.

\subsection{Tensor Algebra}

A tensor is defined by two properties: its type (real, integer, Boolean, etc.) and its shape \citep{rabanser17}. The shape of a tensor consists of its rank (number of indices) and its size (number of elements) along each index. For example, a video can be represented by an integer tensor of shape $(t, x, y, c)$, where $t$ is the number of frames, $x$ and $y$ are a frame's width and height in pixels, and $c$ is the number of color channels (typically 3). A matrix is a rank-2 tensor, a vector a rank-1 tensor, and a scalar a rank-0 tensor. A tensor of rank $r$ and size $n_i$ in the $i$th dimension contains a total of $\prod_{i=1}^r n_i$ elements. The element of a tensor $A$ at position $i_1$ along dimension 1, position $i_d$ along dimension $d$, etc., is denoted by $A_{i_1, \ldots, i_d, \ldots, i_r}$. This generic element of a tensor is often used to represent the tensor itself. The {\em sum} of two tensors $A$ and $B$ with the same shape is a tensor $C$ such that
\[ C_{i_1, \ldots, i_d, \ldots, i_r} = A_{i_1, \ldots, i_d, \ldots, i_r} + B_{i_1, \ldots, i_d, \ldots, i_r}. \]
The {\em tensor product} of two tensors $A$ and $B$ of rank respectively $r$ and $r'$ is a tensor $C$ of rank $r + r'$ such that
\[ C_{i_1, \ldots, i_d, \ldots, i_r, j_1, \ldots, j_{d'}, \ldots, j_{r'}} = A_{i_1, \ldots, i_d, \ldots, i_r} B_{j_1, \ldots, j_{d'}, \ldots, j_{r'}}. \]

Einstein notation simplifies tensor equations by omitting all summation signs and implicitly summing over all repeated indices. For example, $A_{ij} B_{jk}$ represents the product of the matrices $A$ and $B$, summing over $j$ and resulting in a matrix with indices $i$ and $k$:
\[ C_{ik} = A_{ij} B_{jk} = \sum_j A_{ij} B_{jk}. \]
More generally, the {\em Einstein sum} (or einsum for short) of two tensors $A$ and $B$ with common indices $\beta$ is a tensor $C$ such that
\[ C_{\alpha\gamma} = \sum_\beta A_{\alpha\beta} B_{\beta\gamma}, \]
where $\alpha$, $\beta$ and $\gamma$ are sets of indices, $\alpha$ is the subset of $A$'s indices not appearing in $B$, the elements of $\alpha$ and $\beta$ may be interspersed in any order, and similarly for $B$ and $\gamma$. Essentially all linear and multilinear operations in neural networks can be concisely expressed as einsums \citep{rocktaschel18,rogozhnikov22}.

Like matrices, tensors can be decomposed into products of smaller tensors. In particular, the {\em Tucker decomposition} decomposes a tensor into a more compact {\em core tensor} of the same rank and $k$ {\em factor matrices}, each expanding an index of the core tensor into an index of the original one. For example, if $A$ is a rank-3 tensor, in Einstein notation its Tucker decomposition is
\[ A_{ijk} = M_{ip} M'_{jq} M''_{kr} C_{pqr}, \]
where $C$ is the core tensor and the $M$'s are the factor matrices.

\section{Tensor Logic}

\subsection{Representation}
\label{repr}

Tensor logic is based on the answers to two key questions: What is the relation between tensors and relations? And what is the relation between Datalog rules and einsums?

The answer to the first question is that a relation is a compact representation of a sparse Boolean tensor. For example, a social network can be represented by the neighborhood matrix $M_{ij}$, where $i$ and $j$ range over individuals and $M_{ij} = 1$ if $i$ and $j$ are neighbors and 0 otherwise. But for large networks this is an inefficient representation, since almost all elements will be 0. The network can be more compactly represented by a relation, with a tuple for each pair of neighbors; pairs not in the relation are assumed to not be neighbors. More generally, a sparse Boolean tensor of rank $n$ can be compactly represented by an $n$-ary relation with a tuple for each nonzero element, and the efficiency gain will typically increase exponentially with $n$.

The answer to the second question is that a Datalog rule is an einsum over Boolean tensors, with a step function applied elementwise to the result. (Specifically, the Heaviside step function, $H(x) = 1$ if $x > 0$ and 0 otherwise.) For example, consider the rule
\[ {\tt Aunt(x, z)} \leftarrow {\tt Sister(x, y)}, {\tt Parent(y, z)}. \] Viewing the relations ${\tt Aunt(x, z)}$, ${\tt Sister(x, y)}$ and ${\tt Parent(y, z)}$ as the Boolean matrices $A_{xz}$, $S_{xy}$ and $P_{yz}$, respectively,
\[ A_{xz} = H(S_{xy} P_{yz}) = H\left(\sum_y S_{xy} P_{yz}\right) \]
will be 1 iff $S_{xy}$ and $P_{yz}$ are both 1 for at least one $y$. In other words, the einsum $S_{xy} P_{yz}$ implements the join of ${\tt Sister(x, y)}$ and ${\tt Parent(y, z)}$. If $x$ is $z$'s aunt, $y$ is the sibling of $x$ who is also a parent of $z$. The step function is necessary because in general for a given $(x,z)$ pair there may be more than one $y$ for which $S_{xy} = P_{yz} = 1$, leading to a result greater than 1. The step function then reduces this to 1.

Let $U$ and $V$ be arbitrary tensors, and $\alpha$, $\beta$ and $\gamma$ be sets of indices. Then $T_{\alpha\gamma} = H(U_{\alpha\beta} V_{\beta\gamma})$ is a Boolean tensor whose element with indices $\alpha\gamma$ is 1 when there exists some $\beta$ for which $U_{\alpha\beta} V_{\beta\gamma} = 1.$ In other words, $T$ represents the join of the relations corresponding to $U$ and $V$.

Since there is a direct correspondence between tensors and relations and between einsums and Datalog rules, there should also be tensor operations that directly correspond to database join and projection. We are thus led to define tensor projection and tensor join as follows.

The {\em projection} of a tensor $T$ onto a subset of its indices $\alpha$ is
\[ \pi_\alpha(T) = \sum_\beta T_{\alpha\beta}, \]
where $\beta$ is the set of $T$'s indices not in $\alpha$. ($\beta$'s elements may be interspersed with $\alpha$'s in any order.) In other words, the projection of $T$ onto $\alpha$ is the sum for each value of $\alpha$ of all the elements of $T$ with that value of $\alpha$. For example, a vector may be projected onto a scalar by summing all its elements, a matrix onto a column vector by summing each row into an element of the vector, a cubic tensor onto any one of its faces and then that face onto one of its edges and then onto a corner, etc. If the tensors are Boolean and the projection is followed by a step function, tensor projection reduces to database projection.

The {\em join} of two tensors $U$ and $V$ along a common set of indices $\beta$ is
\[ (U \Join V)_{\alpha\beta\gamma} = U_{\alpha\beta} V_{\beta\gamma}, \]
where $\alpha$ is the subset of $U$'s dimensions not in $V$ and similarly for $\gamma$ and $V$. (Again, $\alpha$, $\beta$ and $\gamma$ may be interspersed in any order.) In other words, the join of two tensors on a common subset of indices $\beta$ has one element for each pair of elements with the same value of $\beta$, and that element is their product. If $U$ has rank $r$, $V$ has rank $r'$, and $|\beta| = q$, $U \Join V$ has rank $r + r' - q$. When two tensors have no indices in common, their join reduces to their tensor product (Kronecker product for matrices). When they have all dimensions in common, it reduces to their elementwise product (Hadamard product for matrices). If the tensors are Boolean, tensor join reduces to database join.

A {\em tensor logic program} is a set of tensor equations. The left-hand side (LHS) of a tensor equation is the tensor being computed. The right-hand side (RHS) is a series of tensor joins followed by a tensor projection, and an optional univariate nonlinearity applied elementwise to the result. A tensor is denoted by its name followed by a list of indices, comma-separated and enclosed in square brackets. The join signs are left implicit, and the projection is onto the indices on the LHS. For example, a single-layer perceptron is implemented by the tensor equation
\[ {\tt Y = step(W[i] \, X[i])}, \]
where joining on ${\tt i}$ and projecting it out implements the dot product of ${\tt W}$ and ${\tt X}$. Tensors can also be specified by listing their elements, e.g., ${\tt W = [0.2, 1.9, -0.7, 3]}$ and ${\tt X = [0, 1, 1, 0]}$. Typing ${\tt Y?}$ then causes $Y$ to be evaluated.

Notice that, like the einsum implementations in NumPy, PyTorch, etc., a tensor equation is more general than the original Einstein notation: the summed-over indices are those that do not appear in the LHS, and thus a repeated index may or may not be summed over. For example, the index ${\tt i}$ in
\[ {\tt Y[i] = step(W[i] \, X[i])} \]
is not summed over. The implementation of a multilayer perceptron below utilizes this.

Tensor elements are 0 by default, and equations with the same LHS are implicitly summed. This both preserves the correspondence with logic programming and makes tensor logic programs shorter. Tensor types may be declared or inferred. Setting a tensor equal to a file reads the file into the tensor. Reading a text file results in a Boolean matrix whose $ij$th element is 1 if the $i$th position in the text contains the $j$th word in the vocabulary. (The matrix is not stored in this inefficient form, of course; more on this later.) For example, if the file is the string ``Alice loves Bob'' and it's read into the matrix ${\tt M}$, the result is ${\tt M[0,Alice]} = {\tt M[1,loves]} = {\tt M[2,Bob]} = 1$ and ${\tt M[i,j]} = 0$ for all other ${\tt i}$, ${\tt j}$. (Notice how arbitrary constants, not just integers, can be used as indices.) Conversely, setting a file equal to a tensor writes the tensor to the file.

This is the entire definition of tensor logic. There are no keywords, other constructs, etc. However, it is convenient to allow some syntactic sugar that, while not increasing the expressiveness of the language, makes it more convenient to write common programs. For example, we may allow: multiple terms in one equation (e.g., ${\tt Y = step(W[i] \, X[i] + C)}$); index functions (e.g., ${\tt X[i,t\!+\!1] = W[i,j] \, X[j,t]}$); normalization (e.g., ${\tt Y[i] = softmax(X[i])}$); other tensor functions (e.g., ${\tt Y[k] = concat(X[i,j])}$); alternate projection operators (e.g., ${\tt max\!=}$ or ${\tt avg\!=}$ instead of ${\tt +\!=}$, which ${\tt =}$ defaults to); slices (e.g., ${\tt X[4:8]}$); and procedural attachment (predefined or externally defined functions). Tensor logic accepts Datalog syntax; denoting a tensor with parentheses instead of square brackets implies that it's Boolean. In particular, a sparse Boolean tensor may be written more compactly as a set of facts. For example, the vector ${\tt X = [0, 1, 1, 0]}$ can also be written as ${\tt X(1)}$, ${\tt X(2)}$, with ${\tt X(0)}$ and ${\tt X(3)}$ being implicitly 0. Similarly, reading the string ``Alice loves Bob'' into the matrix ${\tt M}$ produces the facts ${\tt M(0,Alice)}$, ${\tt M(1,loves)}$ and ${\tt M(2,Bob})$.)

As another simple example, a multilayer perceptron can be implemented by the equation
\[ {\tt X[i,j] = sig(W[i,j,k] \, X[i\!-\!1,k])}, \]
where ${\tt i}$ ranges over layers and ${\tt j}$ and ${\tt k}$ over units, and ${\tt sig()}$ is the sigmoid function. Different layers may be of different sizes (and the corresponding weight matrices are implicitly padded with zeros to make up the full tensor). Alternatively, we may use a different equation for each layer.

A basic recursive neural network (RNN) can be implemented by
\[ {\tt X[i,*t\!+\!1] = sig(W[i,j] \, X[j,*t] + V[i,j] \, U[j,t])}, \]
where ${\tt X}$ is the state, ${\tt U}$ is the input, ${\tt i}$ and ${\tt j}$ range over units, and ${\tt t}$ ranges over time steps. The ${\tt *t}$ notation indicates that ${\tt t}$ is a virtual index: no memory is allocated for it, and successive values of the ${\tt X[i]}$ vector are written to the same location. Since RNNs are Turing-complete \citep{siegelmann95}, the implementation above implies that so is tensor logic.

\subsection{Inference}

Inference in tensor logic is carried out using tensor generalizations of forward and backward chaining.

In forward chaining, a tensor logic program is treated as linear code. The tensor equations are executed in turn, each one computing the tensor elements for which the necessary inputs are available; this is repeated until no new elements can be computed or a stopping criterion is satisfied.

In backward chaining, each tensor equation is treated as a function. The query is the top-level call, and each equation calls the equations for the tensors on its RHS until all the relevant elements are available in the data or there are no equations for the subqueries. In the latter case (sub)query elements are assigned 0 by default.

The choice of whether to use forward or backward chaining depends on the application.

\subsection{Learning}

Because there is only one type of statement in tensor logic---the tensor equation---automatic\-ally differentiating a tensor logic program is particularly simple. Univariate nonlinearity aside, the derivative of the LHS of a tensor equation with respect to a tensor on the RHS is just the product of the other tensors on the RHS. More precisely, if
\[ {\tt Y[...] = T[...] \, X_1[...] \ldots X_n[...]}, \]
then
\[ {\tt \frac{\partial Y[...]}{\partial T[...]} = X_1[...] \ldots X_n[...]}. \]
Special cases of this include: if ${\tt Y = AX}$, then ${\tt \partial Y/ \partial X = A}$; if ${\tt Y = W[i] \, X[i]}$, then ${\tt \partial Y/ \partial W[i] = X[i]}$; and if ${\tt Y[i,j] = M[i,k] \, X[k,j]}$, then ${\tt \partial Y[i,j] / \partial M[i,k] = X[k,j]}$.

As a result, the gradient of a tensor logic program is also a tensor logic program, with one equation per equation and tensor on its RHS. Omitting indices for brevity, the derivative of the loss ${\tt L}$ with respect to a tensor ${\tt T}$ is
\[ {\tt \frac{\partial L}{\partial T} = \sum_{E} \, \frac{dL}{dY} \; \frac{dY}{dU} \, \prod_{U \setminus T} \, X}, \]
where ${\tt E}$ are the equations whose RHSs ${\tt T}$ appears in, ${\tt Y}$ is the equation's LHS, ${\tt U}$ is its nonlinearity's argument, and ${\tt X}$ are the tensors in ${\tt U}$.

Learning a tensor logic program requires specifying the loss function and the tensors it applies to by means of one or more tensor equations. For example, to learn an MLP by minimizing squared loss on the last layer's outputs we can use the equation
\[ {\tt Loss = (Y[e] - X[*e,N,j])^2}, \]
where ${\tt e}$ ranges over training examples and ${\tt j}$ over units, ${\tt Y}$ contains the target values, ${\tt X}$ is the MLP as defined above extended with a virtual index for examples, and ${\tt N}$ is the number of layers. By default, all tensors that are not supplied as training data will be learned, but the user can specify if any should remain constant (e.g., hyperparameters). The optimizer itself can be encoded in tensor logic, but typically a pre-supplied one will be used.

While backpropagation in traditional neural networks is applied to the same architecture for all training examples, in tensor logic the structure may effectively vary from example to example, since different equations may apply to different examples, and backpropagating through the union of all possible derivations of the example would be wasteful. Fortunately, a solution to this problem is already available in the form of backpropagation through structure, which for each example updates each equation's parameters once for each time it appears in the example's derivation \citep{goller96}. Applying this to RNNs yields the special case of backpropagation through time \citep{werbos90}.

Learning a tensor logic program consisting of a fixed set of equations is quite flexible, since an equation can represent any set of rules with the same join structure. (E.g., an MLP can represent any set of propositional rules.) Further, tensor decomposition in tensor logic is effectively a generalization of predicate invention. For example, if the program to be learned is the equation
\[ {\tt A]i,j,k] = M[i,p] \, M'[j,q] \, M''[k,r] \, C[p,q,r]} \]
and ${\tt A}$ is the sole data tensor, the learned ${\tt M}$, ${\tt M'}$, ${\tt M''}$ and ${\tt C}$ form a Tucker decomposition of ${\tt A}$; and thresholding them into Booleans turns them into invented predicates.

\section{Implementing AI Paradigms}

The implementations below use forward chaining unless otherwise specified.

\subsection{Neural Networks}

A convolutional neural network is an MLP with convolutional and pooling layers \citep{lecun98}. A convolutional layer applies a filter at every location in an image, and can be implemented by a tensor equation of the form
\[ {\tt Features[x,y] = relu(Filter[dx,dy,ch] \, Image[x\!+\!dx,y\!+\!dy,ch])}, \]
where ${\tt x}$ and ${\tt y}$ are pixel coordinates, ${\tt dx}$ and ${\tt dy}$ are filter coordinates, and ${\tt ch}$ is the RGB channel. A pooling layer combines a block of nearby filters into one, and can be implemented by
\[ {\tt Pooled[x/S,y/S] = Features[x,y]}, \]
where ${\tt /}$ is integer division and ${\tt S}$ is the stride. This results in the filter outputs at ${\tt S}$ successive positions in each dimension being summed into one. This implements sum-pooling; max-pooling would replace ${\tt =}$ with ${\tt max\!=}$, etc. A convolutional and pooling layer can be combined into one with the equation ${\tt Pooled[x/S,y/S] = relu(\ldots)}$.

Graph neural networks (GNNs) apply deep learning to graph-structured data (e.g., social networks, molecules, metabolic networks, the Web) \citep{liu22}. Table~\ref{gnn} shows the implementation of a simple GNN. The network's graph structure is defined by the ${\tt Neig(x,y)}$ relation, with one fact for each adjacent ${\tt (x,y)}$ pair; or equivalently, by the Boolean tensor ${\tt Neig[x,y]} = 1$ if ${\tt x}$ and ${\tt x}$ are adjacent and 0 otherwise. The main tensor is ${\tt Emb[n,l,d]}$, containing the ${\tt d}$-dimensional embedding of each node ${\tt n}$ in each layer ${\tt l}$. Initialization sets each node's 0th-layer embeddings to its features (externally defined or learned). The network then carries out ${\tt L}$ iterations of message passing, one per layer. Each iteration starts by applying one or more perceptron layers to each node. (Table~\ref{gnn} shows one. To preserve permutation invariance, the weights ${\tt W_P}$ do not depend on the node. Although there are no sub/superscripts in tensor logic, I will use them here for brevity.) The GNN then aggregates each node's neighbors' new features ${\tt Z}$ by joining the tensors ${\tt Neig(n,n')}$ and ${\tt Z[n',l,d]}$. For each node, this zeroes out the contributions of all non-neighbors; the result is the sum of the neighbors' features. (Internally, this can be done efficiently by iterating over the node's neighbors or by other methods; see Section~\ref{scaling}.) The aggregated features may then be passed through another MLP (not shown), after which they are combined with the node's features using weights ${\tt W_{Agg}}$ and ${\tt W_{Self}}$ to produce the next layer's embeddings.

\begin{table}
\begin{center}
\caption{Graph neural networks in tensor logic}
\label{gnn}
\smallskip
\begin{tabular}{ll}
\hline
Component & Equation \\
\hline
Graph structure & ${\tt Neig(x,y)}$ \\
Initialization & ${\tt Emb[n,0,d] = X[n,d]}$ \\
MLP & ${\tt Z[n,l,d'] = relu(W_P[l,d',d] \, Emb[n,l,d])}$, etc. \\
Aggregation & ${\tt Agg[n,l,d] = Neig(n,n') \, Z[n',l,d]}$ \\
Update & ${\tt Emb[n,l\!+\!1,d] = relu(W_{Agg} \, Agg[n,l,d] + W_{Self} \, Emb[n,l,d])}$ \\
Node classification & ${\tt Y[n] = sig(W_{Out}[d] \, Emb[n,L,d])}$ \\
Edge prediction & ${\tt Y[n,n'] = sig(Emb[n,L,d] \, Emb[n',L,d])}$ \\
Graph classification & ${\tt Y = sig(W_{Out}[d] \, Emb[n,L,d])}$ \\
\hline
\end{tabular}
\end{center}
\end{table}

The most common applications of GNNs are node classification, edge prediction and graph classification. For two-class problems, each node is classified by doing the dot product of its final embedding with a weight vector, and passing the result through a sigmoid to yield the class probability. For multiclass problems (not shown), each node's final embedding is dotted with a weight vector for each class ${\tt c}$, ${\tt W_{Out}[c,d]}$, yielding a vector of logits that is then passed through a softmax to yield the class probabilities ${\tt Y[n,c]}$. Edge prediction predicts whether there is an edge between each pair of nodes by dotting their embeddings and passing the result through a sigmoid. Graph classification produces a class prediction for the entire graph, and is identical to node classification save for the result being a scalar ${\tt Y}$ instead of a vector ${\tt Y[n]}$.

Attention, the basis of large language models, is also straightforward to implement in tensor logic \citep{vaswani17}. Given an embedding matrix ${\tt X[p,d]}$, where ${\tt p}$ ranges over items (e.g., positions in a text) and ${\tt d}$ over embedding dimensions, the query, key and value matrices are obtained by multiplying ${\tt X}$ by the corresponding weight matrices:
\begin{eqnarray*}
{\tt Query[p,d_k] = W_Q[d_k,d] \, X[p,d]} \\
{\tt Key[p,d_k] =  W_K[d_k,d] \, X[p,d]} \\
{\tt Val[p,d_v] = W_V[d_v,d] \, X[p,d]}
\end{eqnarray*}
Attention can then be computed in two steps, the first of which compares the query at each position with each key:
\[ {\tt Comp[p,p'.] = softmax(Query[p,d_k]) \, Key[p',d_k] \,/\, sqrt(D_k))}, \]
where ${\tt sqrt(D_k)}$ scales the dot products by the square root of the keys' dimension. The notation ${\tt p'.}$ indicates that ${\tt p'}$ is the index to be normalized (i.e., for each ${\tt p}$, softmax is applied to the vector indexed by ${\tt p'}$). The attention head then returns the sum of the value vectors weighted by the corresponding comparisons:
\[ {\tt Attn[p,d_v] = Comp[p,p'] \, Val[p',d_v]}. \]

We can now implement an entire transformer with just a dozen tensor equations (Table~\ref{transf}). As we saw in Subsection~\ref{repr}, a text can be represented by the relation ${\tt X(p,t)}$, stating that the ${\tt p}$th position in the text contains the ${\tt t}$th token. (Tokenization rules are easily expressed in Datalog, and are not shown.) The text's embedding ${\tt EmbX[p,d]}$ is then obtained by multiplying ${\tt X(p,t)}$ by the embedding matrix ${\tt Emb[t,d]}$. The next equation implements positional encoding as in the original paper \citep{vaswani17}; other options are possible. (Incidentally, this equation also shows how conditionals and case statements can be implemented in tensor logic: by joining each expression with the corresponding condition.) The residual stream is then initialized to the sum of the text's embedding and the positional encoding.

\begin{table}
\begin{center}
\caption{Transformers in tensor logic}
\label{transf}
\smallskip
\begin{tabular}{ll}
\hline
Component & Equation(s) \\
\hline
Input & ${\tt X(p,t)}$ \\
Embedding & ${\tt EmbX[p,d] = X(p,t) \, Emb[t,d]}$  \\
Pos. encoding & ${\tt PosEnc[p,d] = Even(d) \, sin(p / L^{d/D_e}) + Odd(d) \, cos(p / L^{d\!-\!1/D_e})}$ \\
Residual stream & ${\tt Stream[0,p,d] = EmbX[p,d] + PosEnc[p,d]}$ \\
Attention & ${\tt Query[b,h,p,d_k] = W_Q[b,h,d_k,d] \, Stream[b,p,d]}$, etc. \\
  & ${\tt Comp[b,h,p,p'.] = softmax(Query[b,h,p,d_k] \, Key[b,h,p',d_k] / sqrt(D_k))}$ \\
  & ${\tt Attn[b,h,p,d_v] = Comp[b,h,p,p'] \, Val[b,h,p',d_v]}$ \\
Merge and & ${\tt Merge[b,p,d_m] = concat(Attn[b,h,p,d_v])}$ \\
~~ layer norm  & ${\tt Stream[b,p,d.] = lnorm(W_S[b,d,d_m] \, Merge[b,p,d_m] + Stream[b,p,d])}$ \\
MLP & ${\tt MLP[b,p] = relu(W_P[p,d] \, Stream[b,p,d])}$, etc. \\
Output & ${\tt Y[p,t.] = softmax(W_O[t,d] \, Stream[B,p,d])}$ \\
\hline
\end{tabular}
\end{center}
\end{table}

Attention is implemented as described above, with two additional indices for each tensor: ${\tt b}$ for the attention block and ${\tt h}$ for the attention head. The attention heads' outputs are then concatenated, added to the residual stream and layer-normalized. MLP layers are implemented as before, with additional indices for block and position, and their outputs are also normalized and added to the stream (not shown). Finally, the output (token probabilities) is obtained by dotting the stream with an output weight vector for each token and passing through a softmax.

\subsection{Symbolic AI}

A Datalog program is a valid tensor logic program. Therefore anything that can be done in Datalog can be done in tensor logic. This suffices to implement many symbolic systems, including reasoning and planning in function-free domains. Accommodating functions (as in Prolog) requires implementing unification in tensor logic \citep{lloyd87}.

\subsection{Kernel Machines}

A kernel machine can be implemented by the equation
\[ {\tt Y[Q] = f(A[i] \, Y[i] \, K[Q,i] + B)}, \]
where ${\tt Q}$ is the query example, ${\tt i}$ ranges over support vectors, and ${\tt f()}$ is the output nonlinearity (e.g., a sigmoid) \citep{scholkopf02}. The kernel ${\tt K}$ is then implemented by its own equation. For example, a polynomial kernel is
\[ {\tt K[i,i'] = (X[i,j] \, X[i',j])^n}, \]
where ${\tt i}$ and ${\tt i'}$ range over examples, ${\tt j}$ ranges over features, and ${\tt n}$ is the degree of the polynomial. A Gaussian kernel is
\[ {\tt K[i,i'] = exp(-(X[i,j] - X[i',j])^2 \,/\, Var)}. \]
(More precisely, ${\tt K}$ is the Gram matrix of the kernel with respect to the examples.) Structured prediction, where the output consists of multiple interrelated elements \citep{bakr07}, can be implemented by an output vector ${\tt Y[Q,k]}$ and equations stating the interactions among outputs and between outputs and inputs.

\subsection{Probabilistic Graphical Models}

A graphical model represents the joint probability distribution of a set of random variables as a normalized product of factors,
\[ P(X\!=\!x) = \frac{1}{Z} \prod_k \phi_k(x_{\{k\}}), \]
where each factor $\phi_k$ is a non-negative function of a subset of the variables $x_{\{k\}}$ and $Z = \sum_x \prod_k \phi_k(x_{\{k\}})$ \citep{koller09}. If each factor is the conditional probability of a variable given its parents (predecessors in some partial ordering), the model is a Bayesian network and $Z = 1$.

Table~\ref{gm} shows how the constructs and operations in discrete graphical models map directly onto those in tensor logic. A factor is a tensor of non-negative real values, with one index per variable and one value of the index per value of the variable. The unnormalized probability of a state $x$ is the product of the element in each tensor corresponding to $x$. A Bayesian network can thus be encoded in tensor logic using one equation per variable, stating the variable's distribution in terms of its conditional probability table (CPT) and the parents' distributions:
\[ {\tt P_X[x] = CPT_X[x,par_1,...,par_n] \, P_1[par_1] \ldots P_n[par_n]}. \]

\begin{table}
\begin{center}
\caption{Graphical models in tensor logic}
\label{gm}
\smallskip
\begin{tabular}{ll}
\hline
Component & Implementation \\
\hline
Factor & Tensor \\
Marginalization & Projection \\
Pointwise product & Join \\
Join tree & Tree-like program \\
P(Query$|$Evidence) & Prog(Q,E)/Prog(E) \\
Belief propagation & Forward chaining \\
Sampling & Selective projection \\
\hline
\end{tabular}
\end{center}
\end{table}

Inference in graphical models is the computation of marginal and conditional probabilities, and consists of combinations of two operations: marginalization and pointwise product. The marginalization of a subset of the variables $Y$ in a factor $\phi$ sums them out, leaving a factor over the remaining variables $X$:
\[ \phi'(X) = \sum_Y \phi(X,Y). \]
Marginalization is just tensor projection. The pointwise product of two potentials over subsets of variables $X$ and $Y$ combines them into a single potential over $X \cup Y$, and is the join of the corresponding tensors.

Every graphical model can be expressed as a join tree, a tree of factors where each factor is a join of factors in the original model. All marginal and conditional queries can be answered in time linear in the size of the tree by successively marginalizing factors and pointwise-multiplying them with the parent's factor. A join tree is a tree-like tensor logic program, i.e., one in which no tensor appears in more than one RHS. As a result, linear-time inference can be carried out by backward chaining over this program. Specifically: the partition function $Z$ can be computed by adding the equation ${\tt Z = T[...]}$ to the program, where ${\tt T[...]}$ is the LHS of the root factor's equation, and querying ${\tt Z}$; the marginal probability of evidence $P(E)$ can be computed by adding $E$ to the program as a set of facts, querying ${\tt Z}$, and dividing by the original one; and the conditional probability of a query given evidence can be computed as $P(E) = P(Q,E)/P(E)$.

However, the join tree may be exponentially larger than the original model, necessitating approximate inference. The two most popular methods are loopy belief propagation and Monte Carlo sampling. Loopy belief propagation is forward chaining on the tensor logic program representing the model. Sampling can be implemented by backward chaining with selective projection (i.e., replacing a projection by a random subset of its terms).

\section{Reasoning in Embedding Space}

The most interesting feature of tensor logic is the new models it suggests. In this section I show how to perform knowledge representation and reasoning in embedding space, and point out the reliability and transparency of this approach.

Consider first the case where an object's embedding is a random unit vector. The embeddings can be stored in a matrix ${\tt Emb[x,d]}$, where ${\tt x}$ ranges over objects and ${\tt d}$ over embedding dimensions. Multiplying ${\tt Emb[x,d]}$ by a one-hot vector ${\tt V[x]}$ then retrieves the corresponding object's embedding. If ${\tt V[x]}$ is a multi-hot vector representing a set,
\[ {\tt S[d] = V[x] \, Emb[x,d]} \]
is the superposition of the embeddings of the objects in the set. The dot product
\[ {\tt D[A] = S[d] \; Emb[A,d]} \]
for some object ${\tt A}$ is then approximately 1 if ${\tt A}$ is in the set and approximately 0 otherwise (with standard deviation $\sqrt{N/D}$, where $N$ is the cardinality of the set and $D$ is the embedding dimension). Thresholding this at $\frac{1}{2}$ then tells us if ${\tt A}$ is in the set with an error probability that decreases with the embedding dimension. This is similar to a Bloom filter \citep{bloom70}.

The same scheme can be extended to embedding a relation. Consider a binary relation ${\tt R(x,y)}$ for simplicity. Then
\[ {\tt EmbR[i,j] = R(x,y) \, Emb[x,i] \, Emb[y,j]} \]
is the superposition of the embeddings of the tuples in the relation, where the embedding of a tuple is the tensor product of the embeddings of its arguments. This is a type of tensor product representation \citep{smolensky90}. It can be computed in time linear in ${\tt |R|}$ by iterating over the tuples adding the corresponding tensor product to the result. The equation
\[ {\tt D[A,B] = EmbR[i,j] \, Emb[A,i] \, Emb[B,j]} \]
retrieves ${\tt R(A,B)}$, i.e., ${\tt D[A,B]}$ is approximately 1 if the tuple ${\tt (A,B)}$ is in the relation and 0 otherwise, since
\begin{eqnarray*}
{\tt D[A,B]} &=& {\tt EmbR[i,j] \, Emb[A,i] \, Emb[B,j]} \\
  & = & {\tt (R(x,y) \, Emb[x,i] \, Emb[y,j]) \, Emb[A,i] \, Emb[B,j]} \\
  & = & {\tt R(x,y) \, (Emb[x,i] \, Emb[A,i]) \, (Emb[y,j] \, Emb[B,j])} \\
  & \simeq & {\tt R(A,B)}.
\end{eqnarray*}       
The penultimate step is valid because einsums are commutative and associative. (In particular, the result does not depend on the order the tensors appear in, only on their index structure.) The last step is valid because the dot product of two random unit vectors is approximately 0.

By the same reasoning, the equation
\[ {\tt D[A,y] = EmbR[i,j] \, Emb[A,i] \, Emb[y,j]} \]
returns the superposition of the embeddings of the objects that are in relation ${\tt R}$ to ${\tt A}$, and
\[ {\tt D[x,y] = EmbR[i,j] \, Emb[x,i] \, Emb[y,j]} \]
returns the entire relation ${\tt R(x,y)}$. ${\tt  EmbR[i,j]}$, ${\tt Emb[x,i]}$ and ${\tt Emb[y,j]}$ form a Tucker decomposition of the data tensor ${\tt D[x,y]}$, with ${\tt  EmbR[i,j]}$ as the core tensor and ${\tt Emb[x,i]}$ and ${\tt Emb[y,j]}$ as the factor matrices.

The relation symbols themselves may be embedded. (E.g., ${\tt R}$, ${\tt A}$ and ${\tt B}$ in ${\tt R(A,B)}$ may all be embedded.) This results in a rank-3 tensor. Relations of arbitrary arity can be reduced to sets of {\em (relation, argument, value)} triples. Thus an entire database can be embedded as a single rank-3 tensor.

The next step is to embed rules. We can embed a Datalog rule by replacing its antecedents and consequents by their embeddings: if the rule is
\[ {\tt Cons(...) \leftarrow Ant_1(...), \ldots, Ant_n(...)}, \]
its embedding is
\[ {\tt EmbCons[...] = EmbAnt_1[...] \ldots EmbAnt_n[...]}, \]
where
\[ {\tt EmbAnt_1[...] = Ant_1(...) \, Emb[...] \ldots Emb[...]}, \]
etc. Reasoning in embedding space can now be carried out by forward or backward chaining over the embedded rules. The answer to a query can be extracted by joining its tensor with its arguments' embeddings, as shown above for any relation. This gives approximately the correct result because each inferred tensor can be expressed as a sum of projections of joins of embedded relations, and the product ${\tt Emb[x,i] \, Emb[x',i]}$ for each of its arguments is approximately the identity matrix. The error probability decreases with the embedding dimension, as before. To further reduce it, we can extract, threshold and re-embed the inferred tensors at regular intervals (in the limit, after each rule application).

The most interesting case, however, is when objects' embeddings are learned. The product of the embedding matrix and its transpose,
\[ {\tt Sim[x,x'] = Emb[x,d] \, Emb[x',d]}, \]
is now the Gram matrix measuring the similarity of each pair of objects by the dot product of their embeddings. Similar objects ``borrow'' inferences from each other, with weight proportional to their similarity. This leads to a powerful form of analogical reasoning that explicitly combines similarity and compositionality in a deep architecture.

If we apply a sigmoid to each equation,
\[ \sigma(x,T) = \frac{1}{(1+e^{-x/T})}, \]
setting its temperature parameter $T$ to 0 effectively reduces the Gram matrix to the identity matrix, making the program's reasoning purely deductive. This contrasts with LLMs, which may hallucinate even at $T=0$. It's also exponentially more powerful than retrieval-augmented generation \citep{jiang25}, since it effectively retrieves the deductive closure of the facts under the rules rather than just the facts.

Increasing the temperature makes reasoning increasingly analogical, with examples that are less and less similar borrowing inferences from each other. The optimal $T$ will depend on the application, and can be different for different rules (e.g., some rules may be mathematical truths and have $T = 0$, while others may serve to accumulate weak evidence and have a high $T$).

The inferred tensors can be extracted at any point during inference. This makes reasoning highly transparent, in contrast with LLM-based reasoning models. It's also highly reliable and immune to hallucinations at sufficiently low temperature, again in contrast with LLM-based models. At the same time, it has the generalization and analogical abilities of reasoning in embedding space. This could make it ideal for a wide range of applications.

\section{Scaling Up}
\label{scaling}

For large-scale learning and inference, equations involving dense tensors can be directly implemented on GPUs. Operations on sparse and mixed tensors can be implemented using (at least) one of two approaches.

The first is separation of concerns: operations on dense (sub)tensors are implemented on GPUs, and operations on sparse (sub)tensors are implemented using a database query engine, by treating (sub)tensors as relations. The full panoply of query optimization can then be applied to combining these sparse (sub)tensors. An entire dense subtensor may be treated as single tuple by the database engine, with an argument pointing to the subtensor. Dense subtensors are then joined and projected using GPUs.

The second and more interesting approach is to carry out all operations on GPUs, first converting the sparse tensors into dense ones via Tucker decomposition. This is exponentially more efficient than operating directly on the sparse tensors, and as we saw in the previous section, even a random decomposition will suffice. The price is that there will be a small probability of error, but this can be controlled by appropriately setting the embedding dimension and denoising results by passing them through step functions. Scaling up via Tucker decompositions has the significant advantage that it combines seamlessly with the learning and reasoning algorithms described in previous sections.

\section{Discussion}

Tensor logic is likely to be useful beyond AI. Scientific computing consists essentially of translating equations into code, and with tensor logic this translation is more direct than with previous languages, often with a one-to-one correspondence between symbols on paper and symbols in code. In scientific computing the relevant equations are then wrapped in logical statements that control their execution. Tensor logic makes this control structure automatically learnable by relaxing the corresponding Boolean tensors to numeric ones, and optionally thresholding the results back into logic. The same approach is applicable in principle to making any program learnable.

Any new programming language faces a steep climb to wide adoption. What are tensor logic's chances of succeeding? AI programming is no longer a niche; tensor logic can ride the AI wave to wide adoption in the same way that Java rode the Internet wave. Backward compatibility with Python is key, and tensor logic lends itself well to it: it can initially be used as a more elegant implementation of einsum and extension of Python to reasoning tasks, and as it develops it can absorb more and more features of NumPy, PyTorch, etc., until it supersedes them.

Above all, adoption of new languages is driven by the big pains they cure and the killer apps they support, and tensor logic very much has these: e.g., it potentially cures the hallucinations and opacity of LLMs, and is the ideal language for reasoning, mathematical and coding models.

Fostering an open-source community around tensor logic will be front and center. Tensor logic lends itself to IDEs that tightly integrate coding, data wrangling, modeling and evaluation, and if it takes off vendors will compete to support it. It is also ideally suited to teaching and learning AI, and this is another vector by which it can spread.

Next steps include implementing tensor logic directly in CUDA, using it in a wide range of applications, developing libraries and extensions, and pursuing the new research directions it makes possible.

For more information on tensor logic, visit tensor-logic.org.

\acks{This research was partly funded by ONR grant N00014-18-1-2826.}

\end{document}